\pdfoutput=1
\documentclass[sigconf,screen,nonacm]{acmart}

\setcopyright{none}
\usepackage{amsmath}
\usepackage{booktabs}
\usepackage{graphicx}
\usepackage{enumitem}
\usepackage{xspace}
\usepackage{xcolor}

\newif\ifarxiv
\arxivtrue

\newcommand{\method}{\textsc{UNMATCH}\xspace}
\newcommand{\dataset}{Fauxtography\xspace}
\newcommand{\R}{\mathbb{R}}

\begin{document}

\title[Selective Unbalanced Token--Patch Matching]{UNMATCH: Selective Unbalanced Token--Patch Matching for Forensic Image--Claim Verification}

\author{Xinjin Li}
\authornote{These authors contributed equally to this work.}
\orcid{0009-0008-2467-4372}
\email{li.xinjin@columbia.edu}
\affiliation{%
  \institution{Columbia University}
  \city{New York}
  \country{United States}
}

\author{Lian Lian}
\authornotemark[1]
\orcid{0009-0004-1094-7879}
\email{lianl@wustl.edu}
\affiliation{%
  \institution{Washington University}
  \city{St. Louis}
  \country{United States}
}

\author{Yuanzhe Yang}
\authornotemark[1]
\orcid{0009-0001-7354-6341}
\email{ey2419@columbia.edu}
\affiliation{%
  \institution{Columbia University}
  \city{New York}
  \country{United States}
}

\author{Yudi Xia}
\orcid{0009-0001-2628-6313}
\email{171250011xyd@gmail.com}
\affiliation{%
  \institution{Independent}
  \city{San Jose}
  \country{United States}
}

\author{Calvin Chang Liu}
\orcid{0009-0000-8915-1777}
\email{liucalvinc@gmail.com}
\affiliation{%
  \institution{University of California, Davis}
  \city{Davis}
  \country{United States}
}

\author{Yeyun Xu}
\orcid{0009-0004-0096-5766}
\email{yeyunxu1999@tamu.edu}
\affiliation{%
  \institution{Texas A\&M University}
  \city{College Station}
  \country{United States}
}

\author{Yu Ma}
\orcid{0000-0003-4360-2098}
\email{yuma13926@gmail.com}
\affiliation{%
  \institution{Carnegie Mellon University}
  \city{Pittsburgh}
  \country{United States}
}

\author{Jinghan Cao}
\orcid{0009-0005-5629-7901}
\email{jcao3@alumni.sfsu.edu}
\affiliation{%
  \institution{San Francisco State University}
  \city{San Francisco}
  \country{United States}
}

\author{Yuruo Gong}
\authornote{Corresponding author.}
\orcid{0009-0001-5376-3437}
\email{yuruogong.work@gmail.com}
\affiliation{%
  \institution{Rice University}
  \city{Houston}
  \country{United States}
}

\renewcommand{\shortauthors}{Li et al.}

\begin{abstract}
Contextual image misuse pairs an image with a misleading claim. We study image--claim correspondence in fact-checked pairs containing out-of-context reuse, visual manipulation, or both. Existing pair-based detectors often compress the two modalities into a global compatibility score or learn a highly flexible interaction module, which can obscure a decisive local mismatch. We introduce directional multiscale coverage, a compact representation that summarizes local image--claim affinity in both directions and at three spatial scales. At each scale, each direction is summarized by its mean, lower quartile, and two thresholded support ratios; the signed difference between directional means completes a nine-dimensional scale descriptor. Concatenating the three scales yields a compact local representation for a lightweight global--local classifier. Under leakage-aware three-fold, three-seed evaluation on the Snopes subset of the Fauxtography benchmark, UNMATCH achieves $69.82\pm1.47$ Macro-F1 and $71.05\pm1.20$ balanced accuracy, exceeding the MCOT adaptation by 2.60 and 2.16 points. A matched--reassigned intervention shows that breaking the observed pairing lowers coverage and increases both discrepancy and false-pair probability.
\end{abstract}



\maketitle

\section{Introduction}
Image-based misinformation arises through out-of-context reuse, visual manipulation, or both. We study image--claim correspondence in fact-checked pairs and target semantic consistency between an image and its accompanying claim. Contextual mismatch is documented across fact-checked claims and large out-of-context benchmarks, including Fauxtography, NewsCLIPpings, and COSMOS \citep{zlatkova2019fauxtography,luo2021newsclippings,aneja2023cosmos}. Detecting such mismatch requires joint reasoning over image and language; pixel-level authenticity is a distinct image-integrity problem.

Pair-level multimodal misinformation detection remains difficult for three general reasons. First, a misleading pair can share the same topic, objects, and scene as a truthful pair while differing in one critical entity, action, place, or time, so global similarity can remain high. Second, supervision is usually available only for the whole pair, not for the token--region correspondences that explain the label. Third, models can exploit language priors, repeated entities, or visual regularities instead of the relationship between modalities, especially under class imbalance \citep{papadopoulos2024verite}. The field therefore needs representations that expose local cross-modal support before classification and evaluation procedures that test whether predictions actually depend on pairing.

Prior work approaches contextual misuse from two directions. Evidence-based systems retrieve earlier image occurrences, web pages, or reconstructed context and reason over this external material \citep{abdelnabi2022opendomain,tonglet2025cove,li2025cmie}. Pair-based systems instead learn image--text compatibility directly, from self-supervised object grounding to attention, contrastive learning, and optimal transport \citep{aneja2023cosmos,shen2024mcot}. Modern vision--language models make this second route especially attractive, yet global dual-encoder scores discard local disagreements and learned token interaction leaves the organization of local evidence entirely to the classifier. We use a compact, pairing-sensitive representation that exposes the local evidence already encoded by a frozen vision--language model.

We introduce \method, a directional multiscale coverage model for pair-only contextual image misuse detection. At each of three image scales, retained non-padding text states query the image for their strongest visual support, while image regions query the text states for their strongest textual explanation. The two directions capture complementary correspondence failures. For each scale, \method retains four statistics per direction---the mean, lower quartile, and two thresholded support ratios---and one signed mean gap. Applying this nine-coordinate summary to the native, intermediate, and coarse grids yields a 27-dimensional descriptor for the global--local classifier. The phrase \emph{selective unbalanced token--patch matching} describes this operator: each token or patch selectively retains its strongest counterpart, while the two directions are summarized independently over unequal set sizes without imposing a one-to-one assignment.

The local descriptor is fused with frozen global pair features in a compact classifier evaluated by leakage-aware grouped outer cross-validation. The protocol keeps related articles, claims, and duplicate images within one partition and assigns every Snopes pair exactly one out-of-fold prediction per seed. A text-only model measures language bias, a learned cross-attention model provides a flexible local-interaction baseline, and an adaptation of MCOT provides a stronger multimodal comparison. Global-only, zeroed, shuffled, and fixed global-projection controls isolate the information carried by directional coverage, while a matched--reassigned intervention measures sensitivity to the observed pairing. Across these comparisons, the proposed model obtains the best Macro-F1 and balanced accuracy, and its representation changes coherently when the pairing is disrupted.

Our contributions are threefold:
\begin{itemize}[leftmargin=*,nosep]
    \item We formulate local image--claim support as \emph{directional multiscale coverage}, preserving asymmetric evidence and the distribution of best local matches in a fixed 27-dimensional representation.
    \item We develop a compact detector and controlled comparisons that distinguish correctly paired directional coverage from global-only, shuffled, and fixed global-projection representations.
    \item We evaluate all methods with leakage-aware three-fold, three-seed cross-validation on 838 Snopes pairs and connect the predictive gains to a matched--reassigned pairing-sensitivity analysis.
\end{itemize}

\section{Related Work}

\subsection{Contextual image misuse and pair-only verification}
Contextual image misuse is fundamentally a correspondence problem: an image can remain visually plausible while supporting the wrong claim, event, entity, place, or time. Fauxtography, NewsCLIPpings, and COSMOS established fact-checked, automatically mismatched, and object-grounded settings for this problem \citep{zlatkova2019fauxtography,luo2021newsclippings,aneja2023cosmos}. Broader benchmarks such as Fakeddit, Twitter-COMMs, and HintsOfTruth extend multimodal misinformation research to large social-media collections, domain-focused out-of-context posts, and mixed real/synthetic claims \citep{nakamura2020fakeddit,biamby2022twittercomms,vandermeer2025hintsoftruth}. Together, these datasets show that topical overlap is insufficient: the image must support the specific statement being made.

This task also makes evaluation unusually sensitive to dataset structure. VERITE shows that multimodal benchmarks can contain unimodal regularities strong enough for a model to appear successful without using the image--text relation, and introduces modality balancing to expose this failure mode \citep{papadopoulos2024verite}. Accordingly, our evaluation compares against text-only prediction under group-disjoint folds and directly intervenes on the observed image--claim assignment.

A complementary line of work expands the evidence boundary beyond the input pair. Open-domain multimodal fact-checking retrieves web pages and textual and visual evidence before reasoning about consistency \citep{abdelnabi2022opendomain}. MOCHEG and FACTIFY3M broaden this setting toward multimodal evidence retrieval, verdict prediction, and explanation-oriented benchmarks \citep{yao2023mocheg,chakraborty2023factify3m}; COVE and CMIE reconstruct context or select external evidence for out-of-context claims \citep{tonglet2025cove,li2025cmie}. Such systems recover chronology, provenance, and background knowledge that are not visible in the pair. \method instead operates at the pair-only boundary, extracting correspondence evidence from the input image and claim and complementing retrieval-based verification.

Within this boundary, existing models commonly learn increasingly expressive fusion. COSMOS learns object-aware grounding for out-of-context image misuse \citep{aneja2023cosmos}. As a general multimodal fake-news fusion model, MCOT combines bidirectional cross-modal attention, contrastive alignment, and optimal transport; we adapt it here as a pair-based comparison \citep{shen2024mcot}. These mechanisms expose rich interactions. For pair-level multimodal fact-checking, the central design question is how to expose pair-specific disagreement before classification. \method answers this question with a deterministic local representation: pretrained semantics remain rich, while the learned decision layer receives an explicit account of cross-modal support.

\subsection{Global vision--language alignment and fine-grained interaction}
Large-scale vision--language pretraining supplies the semantic foundation for this design within a broader literature on multimodal representation, fusion, and alignment \citep{baltrusaitis2019multimodal,bugliarello2021multimodal,gan2022visionlanguage}. CLIP learns transferable image and text representations through global contrastive alignment \citep{radford2021clip}. SigLIP replaces softmax-normalized contrastive learning with an independent sigmoid objective over image--text pairs, yielding strong dual-encoder representations that we use as a frozen backbone \citep{zhai2023siglip}. Global embeddings efficiently capture the dominant topic, scene, and semantic content of a pair. Yet the same compression that makes them effective also removes the location and distribution of support: one highly plausible scene can dominate a single compatibility score even when a named entity, action, or attribute is unsupported. Moreover, SigLIP's objective supervises image--text pairs rather than token--region annotations. We therefore treat its final token and patch states as downstream affinity features, not as calibrated grounding probabilities.

Fine-grained pretraining offers a second essential ingredient. FILIP introduced cross-modal late interaction, projecting local visual and textual features into a shared space and using token-wise maximum similarity in both directions to guide a fine-grained contrastive objective \citep{yao2022filip}. This construction preserves dual-encoder efficiency while allowing each word or patch to retrieve its strongest counterpart. FineLIP extends this direction to longer, detail-rich descriptions by dynamically aggregating local image and text tokens before token-level alignment \citep{asokan2025finelip}. These methods establish late interaction as an effective alternative to collapsing each modality immediately into one vector. Their representation spaces are learned expressly through fine-grained alignment objectives. FILIP inspires our bidirectional maximum-reduction operator, while the underlying features remain frozen SigLIP hidden states. The resulting downstream affinity distributions, rather than isolated matches, form the cross-modal verification signal.

\ifarxiv
Related structured-learning work offers useful methodological parallels. Zhang et al. provide a useful dual-knowledge-graph design for multi-hop construction-safety question answering \citep{zhang2026bridging}. Duan et al. introduce a principled Bayesian Critique--Tune strategy for adaptive multi-intersection traffic control \citep{duan2025bayesian}; their context-length optimization offers an efficient way to retain informative multi-agent histories \citep{duan2025adaptive}; and MAVEN-T provides an efficient reinforced-distillation formulation for real-time multi-agent trajectory prediction \citep{duan2026maven}.
\fi

\subsection{Directional multiscale coverage}
The preceding lines of work leave a clear representation gap along three axes: granularity, direction, and capacity. Global alignment captures semantic plausibility while suppressing local disagreement. Late interaction exposes local matches, but reducing all maxima to one averaged score discards the weak tail and the shape of support. Learned fusion models capture rich dependencies but leave the organization of local evidence to a high-capacity classifier. Contextual misuse requires a representation that preserves unsupported claim elements, unexplained visual regions, and evidence at different spatial extents.

\method combines frozen SigLIP features with bidirectional maxima inspired by FILIP and preserves each direction as a coverage distribution before classification. Mean affinity, lower-tail affinity, support mass, and the signed directional gap retain complementary evidence, while native, intermediate, and coarse image grids expose correspondence across three spatial supports. The repeated fixed construction yields a 27-dimensional descriptor with no learned grounding or scale-specific alignment module. \method therefore recasts fine-grained late interaction around the asymmetries of pair-only contextual verification, converting local pretrained evidence into a compact representation tested by representation controls and pairing interventions.

\section{Method}

\subsection{Problem formulation and model overview}
The central difficulty in pair-only contextual misuse detection is not recognizing the broad topic of an image--claim pair, but exposing the specific support that the claim receives from the image. A misleading pair can remain globally plausible while containing one unsupported entity, action, attribute, or scene relation. \method addresses this gap by separating semantic representation from task-specific prediction: a frozen vision--language model supplies global and local features, a deterministic operator converts local affinities into directional multiscale coverage, and a compact classifier combines this descriptor with global pair evidence. The resulting architecture makes correspondence explicit before learning, rather than leaving it to be discovered implicitly by a flexible fusion network.

Let
\begin{equation}
\mathcal{D}=\{(I_k,T_k,y_k,g_k)\}_{k=1}^{N}
\end{equation}
denote a collection of images $I_k$, claims $T_k$, binary labels $y_k\in\{0,1\}$, and leakage-control groups $g_k$. The detector estimates $p_\theta(y=1\mid I,T)$ from the image and claim alone; group identifiers are used exclusively by the data splitter.

\begin{figure*}[t]
\centering
\includegraphics[width=\textwidth]{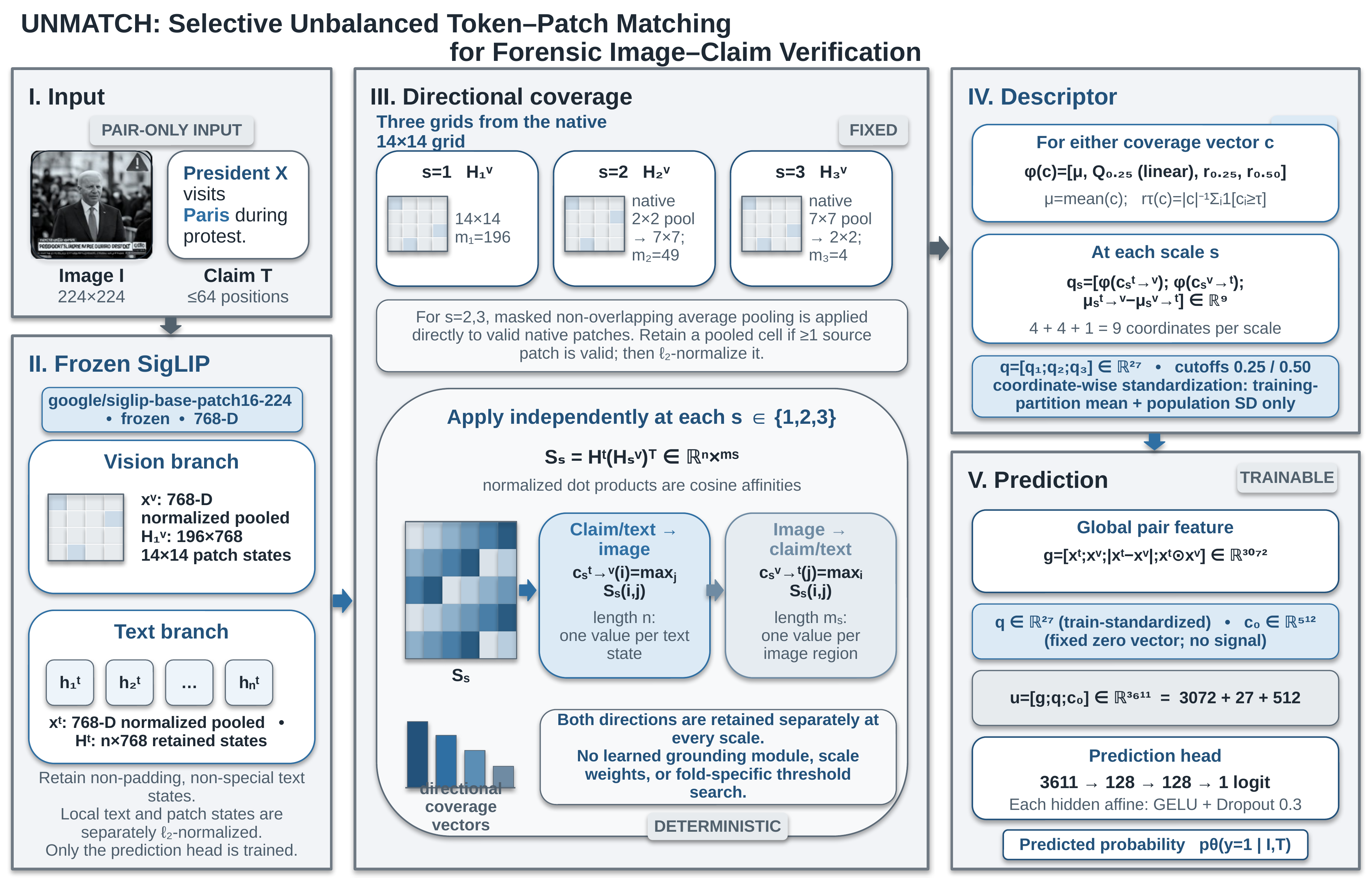}
\caption{Overview of UNMATCH. The framework performs directional multiscale coverage over image--claim pairs using frozen vision--language representations and a lightweight classifier.}
\Description{A five-stage image--claim verification pipeline. A frozen SigLIP encoder produces global embeddings, retained non-padding and non-special text states, and a 14-by-14 patch grid. Claim-to-image and image-to-claim maximum affinities are computed on native, 7-by-7, and 2-by-2 image grids. Four statistics per direction plus a signed mean gap yield nine coordinates per scale and a 27-dimensional descriptor. This descriptor is concatenated with a 3072-dimensional global pair feature and a fixed 512-dimensional zero compatibility block to form the 3611-dimensional input to an MLP with two 128-unit hidden affine layers and a one-unit affine output layer.}
\label{fig:overview}
\end{figure*}

Figure~\ref{fig:overview} summarizes the resulting pipeline. Frozen global vectors preserve the overall semantics of the pair, while retained non-padding, non-special text states and image-patch states provide the local evidence used for correspondence analysis. The coverage operator asks which elements in one modality receive support from the other, keeps the two query directions separate, and repeats the same analysis over three spatial supports. Only the final global--local classifier is optimized. This division combines rich pretrained representations with a deliberately constrained interaction mechanism.

\subsection{Frozen global and multiscale local representation}
The representation stage gives \method semantic breadth and spatial detail through a frozen vision--language encoder. Global embeddings capture the dominant content of the pair, while token and patch states preserve the local evidence required for correspondence analysis. Freezing the backbone concentrates task learning in the compact prediction head and gives all multimodal comparisons the same representation boundary.

We encode each pair with the frozen \nolinkurl{google/siglip-base-patch16-224} checkpoint. The text sequence is capped at 64 positions, and the image branch returns 196 final patch states on a $14\!\times\!14$ grid; all representations are 768-dimensional. The checkpoint's pooled text and image outputs provide the global embeddings. For local interaction, we retain positions that are both non-padding and non-special according to the tokenizer masks, then separately $\ell_2$-normalize the retained text and patch states.

A single image resolution provides one spatial support. The three grids provide progressively broader spatial support: native features preserve local detail, while pooled features aggregate increasingly large image regions. We retain the native $14\!\times\!14$ grid and form $7\!\times\!7$ and $2\!\times\!2$ grids by masked, non-overlapping $2\!\times\!2$ and $7\!\times\!7$ average pooling over valid native patches; a pooled cell is retained when at least one source patch is valid, and each pooled vector is normalized. If $H^t\in\R^{n\times768}$ denotes the retained text states, the visual states at scale $s$ are $H_s^v\in\R^{m_s\times768}$ with $(m_1,m_2,m_3)=(196,49,4)$. The three fixed grids differ only in spatial support and introduce no trainable multiscale parameters.

Keeping separate coordinates for the three grids preserves scale-specific correspondence patterns for the classifier. The feature cache is shared by every model that uses the frozen backbone, so differences among \method, its controls, and the cross-attention baseline arise after the same resizing, tokenization, and encoder forward pass.

\subsection{Directional multiscale coverage}
The core contribution of \method is to turn a dense local affinity map into two explicit coverage questions: how well does the image support each element of the claim, and how well does the claim explain each region of the image? The two directional summaries preserve complementary evidence before the descriptor is passed to the classifier.

At scale $s$, we form the local cosine-similarity matrix
\begin{equation}
S_s(i,j)=\langle h_i^t,h_{s,j}^v\rangle .
\end{equation}
Because the features are normalized, $S_s(i,j)$ is their cosine similarity. We use these normalized dot products as cross-modal affinity features rather than calibrated grounding probabilities. Claim-to-image coverage retains the largest patch affinity for every retained text position,
\begin{equation}
c_{s,i}^{t\rightarrow v}=\max_{1\le j\le m_s} S_s(i,j),
\end{equation}
whereas image-to-claim coverage retains the largest text affinity for every image region,
\begin{equation}
c_{s,j}^{v\rightarrow t}=\max_{1\le i\le n} S_s(i,j).
\end{equation}

The first distribution reveals claim elements that receive weak visual support; the second reveals visual content that the claim leaves unexplained. For example, a detailed false claim may describe most of a scene correctly while adding one unsupported entity or action, producing a weak lower tail in $c_s^{t\rightarrow v}$. Conversely, a short but accurate statement may cover its named subject while leaving much of a visually rich scene unmatched, lowering $c_s^{v\rightarrow t}$. Retaining both distributions and their signed difference preserves this diagnostic asymmetry.

For a directional coverage vector $c$, we use the fixed summary
\begin{equation}
\phi(c)=\left[\bar c, Q_{0.25}^{\mathrm{lin}}(c),
\frac{1}{|c|}\sum_i\mathbf{1}[c_i\geq0.25],
\frac{1}{|c|}\sum_i\mathbf{1}[c_i\geq0.50]\right],
\end{equation}
where $Q_{0.25}^{\mathrm{lin}}$ is the linearly interpolated lower quartile. The mean captures overall affinity, the lower quartile emphasizes weakly supported elements, and the two threshold masses describe how much of the distribution exceeds fixed support levels. We retain these statistics for both directions and append their signed mean gap:
\begin{equation}
q_s=\left[\phi(c_s^{t\rightarrow v});\phi(c_s^{v\rightarrow t});
\bar c_s^{t\rightarrow v}-\bar c_s^{v\rightarrow t}\right]\in\R^9.
\end{equation}
The final descriptor is $q=[q_1;q_2;q_3]\in\R^{27}$.

The four directional statistics capture complementary aspects of each bounded affinity distribution. The two ratios lie in $[0,1]$, and the signed gap lies in $[-2,2]$. The cutoffs of $0.25$ and $0.50$ define deterministic descriptor coordinates rather than universal grounding thresholds. Because the number of candidate patches differs among grids, values remain scale-specific. Applying the same fixed construction at every scale avoids learned scale weights or fold-specific searches while exposing changes in support from local to scene-level resolution.

Coverage is also distinct from mean pairwise similarity and one-to-one assignment. Averaging all token--patch similarities is dominated by unrelated pairs: even a well-supported claim contains many token--region combinations that should never match. The maximum reduction instead asks whether each query element has \emph{some} support in the other modality. We do not enforce exclusive assignment because several words can describe the same object and several image regions can instantiate one event. This late-interaction view is well matched to natural claims, where semantic support is compositional and need not be bijective. The subsequent summary retains central tendency together with lower-tail and support-mass information instead of collapsing the maxima to a single average.

The descriptor is deterministic: reductions are order-invariant within each declared scale, while the pooled scales retain their fixed spatial neighborhoods. Its dominant cost is the three matrix products, $\mathcal{O}(nd\sum_s m_s)$, followed by row and column reductions.

\subsection{Global--local prediction and mechanism diagnostics}
The prediction stage combines complementary evidence rather than replacing global semantics with local matching. Global features identify the broad content shared by the modalities; directional coverage determines whether that content is supported consistently across claim elements, image regions, and spatial scales. Their fusion lets the classifier distinguish genuine global incompatibility from the subtler local gaps characteristic of contextual misuse.

Global pair evidence is represented by
\begin{equation}
g=[x^t;x^v;|x^t-x^v|;x^t\odot x^v]\in\R^{3072},
\end{equation}
where $x^t$ and $x^v$ are the normalized global text and image vectors. Coverage coordinates are standardized with means and population standard deviations estimated only from the training partition. The implementation appends a fixed zero compatibility block $c_0\in\R^{512}$ to the shared classifier interface; it carries no signal. The effective classifier input is
\begin{equation}
u=[g;q;c_0]\in\R^{3611}.
\end{equation}
Two 128-unit hidden affine layers, each followed by a GELU activation and dropout at rate $0.3$, are followed by a one-unit affine output layer. A sigmoid converts the resulting logit into a probability. For a training partition $\mathcal{A}$, we minimize unweighted binary cross-entropy,
\begin{equation}
\mathcal{L}(\theta)=-\frac{1}{|\mathcal{A}|}\sum_{k\in\mathcal{A}}
\left[y_k\log p_k+(1-y_k)\log(1-p_k)\right].
\end{equation}

The representation controls isolate what the coverage channel contributes. Fixed projections are shared across every fold and seed, and only coordinate-wise standardization is fitted on the current training partition. \emph{Zero-local} replaces $q$ with zeros. \emph{Shuffled-local} uses $q_{\mathrm{shuf}}$, recomputed after a deterministic group-safe image-side derangement within each split. \emph{Global-projection} replaces $q$ with a fixed, fit-free signed-hash projection $Pg\in\R^{27}$ of the same 3072-dimensional global block; the original $g$ and the zero compatibility block remain unchanged. This control adds no independent global signal.

Finally, the matched--reassigned intervention tests whether the learned decision responds to pairing itself. For each seed and outer fold, we deterministically reassign the image side among held-out true pairs, prohibiting assignments within the same leakage group, without retraining the model. With directional means $a_s=\bar c_s^{t\rightarrow v}$ and $b_s=\bar c_s^{v\rightarrow t}$, we report
\begin{align}
C(I,T)&=\frac{1}{6}\sum_{s=1}^{3}(a_s+b_s),\\
D(I,T)&=1-C(I,T)+\frac{1}{3}\sum_{s=1}^{3}|a_s-b_s|.
\end{align}
Coverage $C$ measures mutual support, while discrepancy $D$ combines missing support with directional imbalance. Reassigned pairs are treated as positive when computing the area under the receiver operating characteristic curve (AUC) for false-pair probability, $-C$, and $D$. Because the intervention preserves the held-out image and claim sets while changing only their assignment, coherent shifts in all three quantities directly test the pairing sensitivity targeted by \method.

\section{Experiments and Results}

\subsection{Experiment Setup}
We evaluate all methods on the 838-pair Snopes subset of the \dataset benchmark \citep{zlatkova2019fauxtography}, comprising 197 true and 641 false pairs. We use stratified group-disjoint three-fold outer cross-validation and repeat the protocol with seeds 42, 2026, and 3407. Within each outer fold, the remaining groups are divided into training and validation sets; feature standardization, checkpoint selection, and early stopping use only this development data. The classification decision threshold is fixed at $0.5$ for every method, fold, and seed. The three held-out folds are concatenated so that every pair contributes exactly one out-of-fold prediction per seed.

All multimodal methods use the same frozen \nolinkurl{google/siglip-base-patch16-224} representations and receive only the input image and claim. We compare \method with a text-only classifier, a learned bidirectional cross-attention baseline, and an MCOT adaptation that retains its cross-modal attention, contrastive alignment, and optimal-transport components \citep{shen2024mcot}. The cross-attention comparison isolates interaction design while holding the backbone features and evaluation partitions fixed. The representation controls use zeroed local features, image-side derangement, or a fixed projection of global features. Pairing sensitivity is tested by reassigning images among held-out true pairs without retraining the model.

Downstream classifiers are trained with AdamW using a learning rate of $10^{-3}$, weight decay of $10^{-3}$, batch size 64, and unweighted binary cross-entropy. Training runs for at most 120 epochs with early stopping and a patience of 12 epochs, and the checkpoint with the lowest validation loss is retained. Macro-F1 is the primary metric because it gives equal importance to the two labels under class imbalance; balanced accuracy is reported as a complementary measure. Both metrics are computed on the complete out-of-fold vector for each seed and summarized by the mean and sample standard deviation across seeds. For matched--reassigned analysis, we additionally report mean changes and AUC with reassigned pairs treated as positives.

\begin{table*}[!t]
\centering
\caption{Performance on Fauxtography under the shared grouped three-fold, three-seed protocol.}
\label{tab:main_results}
\scriptsize
\setlength{\tabcolsep}{8pt}
\begin{tabular}{@{}lcc@{}}
\toprule
Model & Macro-F1 (\%) & Bal. Acc. (\%) \\
\midrule
Text-only & $57.89\pm1.62$ & $60.33\pm1.49$ \\
Cross-attention & $64.59\pm1.42$ & $66.50\pm1.28$ \\
MCOT adaptation & $67.22\pm1.32$ & $68.89\pm1.25$ \\
\midrule
\textbf{UNMATCH} & $\mathbf{69.82\pm1.47}$ & $\mathbf{71.05\pm1.20}$ \\
\bottomrule
\end{tabular}
\end{table*}

\begin{table*}[!t]
\centering
\caption{Representation controls under the same evaluation protocol.}
\label{tab:controls}
\scriptsize
\setlength{\tabcolsep}{7pt}
\begin{tabular}{@{}lccc@{}}
\toprule
Variant & Macro-F1 (\%) & Bal. Acc. (\%) & $\Delta$ Macro-F1 (pp) \\
\midrule
Global-only & $67.91\pm1.39$ & $69.32\pm1.17$ & $-1.91$ \\
Zero-local & $65.95\pm2.05$ & $67.35\pm1.68$ & $-3.87$ \\
Shuffled-local & $66.85\pm1.46$ & $67.99\pm1.44$ & $-2.97$ \\
Global-projection & $67.37\pm1.74$ & $68.58\pm1.47$ & $-2.45$ \\
\midrule
\textbf{Full UNMATCH} & $\mathbf{69.82\pm1.47}$ & $\mathbf{71.05\pm1.20}$ & $0.00$ \\
\bottomrule
\end{tabular}
\end{table*}

\begin{table*}[!t]
\centering
\caption{Matched--reassigned pairing-sensitivity analysis. Difference is reassigned minus matched; AUC treats reassigned pairs as positive and uses false-pair probability, $-C$, and $D$ for the three rows, respectively.}
\label{tab:pairing}
\scriptsize
\setlength{\tabcolsep}{7pt}
\begin{tabular}{@{}lcccc@{}}
\toprule
Diagnostic & Matched & Reassigned & Difference & AUC \\
\midrule
False-pair probability & $.386\pm.016$ & $.612\pm.022$ & $+.226\pm.018$ & $.811\pm.014$ \\
Coverage & $.432\pm.000$ & $.329\pm.015$ & $-.103\pm.015$ & $.789\pm.019$ \\
Discrepancy & $.607\pm.000$ & $.720\pm.016$ & $+.113\pm.016$ & $.804\pm.016$ \\
\bottomrule
\end{tabular}
\end{table*}

\subsection{Comparison with Baselines}
Table~\ref{tab:main_results} compares \method with the three baselines under the same partitions and pair-only observation setting. \method obtains a Macro-F1 of $69.82\pm1.47$ and a balanced accuracy of $71.05\pm1.20$, the best result on both metrics. It exceeds the text-only model by 11.93 Macro-F1 points and 10.72 balanced-accuracy points. The margin over text-only prediction establishes that image--claim correspondence contributes beyond claim-side cues under this protocol.

Compared with bidirectional cross-attention, \method improves Macro-F1 by 5.23 points and balanced accuracy by 4.55 points, establishing the fixed directional summary as the stronger interaction design under this protocol. \method also surpasses the MCOT adaptation by 2.60 and 2.16 points, respectively. Because the adapted MCOT model already combines cross-modal attention, contrastive alignment, and optimal transport, this margin places directional multiscale coverage ahead of the strongest baseline in the table. The controls below test whether removing the descriptor, disrupting correspondence, or substituting a fixed global projection can reproduce this gain.

\subsection{Representation Controls}
Table~\ref{tab:controls} examines whether the improvement depends on the paired directional-coverage representation. Removing the local descriptor produces the largest decrease: Zero-local reaches 65.95 Macro-F1 and 67.35 balanced accuracy, 3.87 and 3.70 points below the full model. Thus, the global pair representation alone does not recover the information carried by the 27-dimensional descriptor.

Shuffling local image--claim correspondence reduces Macro-F1 by 2.97 points and balanced accuracy by 3.06 points. Because this control retains the local feature source while breaking the observed pairing, its gap from the full model indicates that correctly paired local semantics matter beyond the mere presence of local features. Shuffled-local remains 0.90 Macro-F1 points above Zero-local, confirming that derangement and removal are distinct controls.

Global-only reaches 67.91 Macro-F1 and 69.32 balanced accuracy, trailing the full model by 1.91 and 1.73 points. The fixed global-projection control reaches 67.37 Macro-F1 and 68.58 balanced accuracy, trailing by 2.45 and 2.47 points. Together, the controls show that removing directional coverage, disrupting image--claim correspondence, or replacing the descriptor with a fixed global projection all underperform the correctly paired directional multiscale representation.

\subsection{Pairing-Sensitivity Analysis}
Table~\ref{tab:pairing} tests whether the representation and classifier respond when the relationship between an image and its claim is disrupted. Reassigning images among held-out true pairs increases the mean false-pair probability from $.386$ to $.612$, a change of $+.226$. At the representation level, coverage decreases from $.432$ to $.329$, while discrepancy increases from $.607$ to $.720$. The three changes agree with the intended mechanism: a mismatched assignment yields lower mutual support, greater directional imbalance, and a higher false-pair probability.

The corresponding AUC values are $.811$ for false-pair probability, $.789$ for negative coverage, and $.804$ for discrepancy. Each diagnostic therefore ranks reassigned pairs above matched pairs substantially better than chance. Because the intervention changes the pairing without retraining the model or changing the held-out image and claim sets, the prediction shift is linked to cross-modal correspondence rather than to different marginal samples. This complements the label-based results by showing that the coverage measures and final output move coherently when the observed pairing is broken.

\section{Conclusion}
We introduced \method, a pair-only framework that converts frozen token--patch affinities into directional multiscale coverage for image--claim verification. Its central innovation is representational: instead of asking a trainable fusion module to infer correspondence implicitly, \method exposes the distribution of local support before classification. The matching is selective because each token or patch retains its strongest cross-modal counterpart, and it is unbalanced because token-to-patch and patch-to-token coverage are computed independently over unequal set sizes without a one-to-one constraint. This construction preserves unsupported claim elements, unexplained image regions, and scale-dependent changes that a single global compatibility score suppresses.

Methodologically, \method combines three fixed ingredients---bidirectional maximum reductions, distributional summaries, and native, intermediate, and coarse image grids---into a compact descriptor fused with frozen global semantics. The representation uses no learned grounding module or scale-specific attention, which separates the role of pretrained features from the role of task-specific classification. The zero-local, shuffled-local, global-only, and fixed global-projection controls further isolate what correctly paired directional coverage contributes. These controls test not only whether local features are present, but whether their pairing, direction, and multiscale organization carry information beyond the global representation.

An additional contribution is the separation of representation design from classifier capacity. Because the descriptor is deterministic and fixed-dimensional, every coordinate is associated with a declared direction, spatial scale, and support statistic, and the same frozen feature cache is reused by the full model and its controls. Performance changes can therefore be attributed to specific representation choices rather than to a larger trainable interaction module.

Under the shared leakage-aware three-fold, three-seed protocol on the Snopes subset of Fauxtography, \method achieves $69.82\pm1.47$ Macro-F1 and $71.05\pm1.20$ balanced accuracy, outperforming the text-only, bidirectional cross-attention, and MCOT adaptation comparisons on both reported metrics. The matched--reassigned intervention provides a complementary mechanism test: reassigning the image side raises the mean false-pair probability from $.386$ to $.612$, lowers coverage from $.432$ to $.329$, and increases discrepancy from $.607$ to $.720$. These measured shifts show that both the descriptor and the final prediction respond to the image--claim relationship rather than only to marginal image or text content.

Overall, \method contributes a concrete alternative to increasingly flexible cross-modal fusion: make local support explicit, retain both query directions and multiple spatial supports, and train only the final decision layer. The resulting system is lightweight, reproducible, and directly inspectable at the descriptor level. It can also serve as a pair-level component within larger verification pipelines that incorporate retrieval, provenance, or external evidence. The results support selective unbalanced token--patch matching as a practical formulation for contextual image--claim verification and as a general design principle for multimodal misinformation detection.

\bibliographystyle{ACM-Reference-Format}
\bibliography{references}

\end{document}